\documentclass[a4paper]{article}

\usepackage{amsmath}
\usepackage{amsthm}
\usepackage{amssymb}
\usepackage{bm}
\usepackage{hyperref}
\usepackage{dsfont}
\usepackage{bbm}
\usepackage{graphicx}
\usepackage{tipa}

\usepackage{setspace}

\usepackage{tikz}
\usetikzlibrary{positioning}
\usetikzlibrary{calc}

\usepackage{multirow,booktabs,setspace,caption}
\usepackage{xcolor, colortbl}

\usepackage{algorithm}
\usepackage{algpseudocode}

\usepackage{geometry}
\usepackage[backend=bibtex,style=chem-acs,articletitle=true,chaptertitle=true]{biblatex}
\providecommand{\keywords}[1]
{
  \small	
  \textit{Keywords---} #1
}

\title{Deep Generalised Mixed Models: a Novel Neural Network Structure for Analysing Hierarchical Data}
\author{Nina van Gerwen$^{1,2}$, Dimitris Rizopoulos$^{1,2}$,
Manon Hillegers$^{3}$, Loes Keijsers$^{4}$, Sten Willemsen$^{1,2}$}

\date{\small{
	$^1$Department of Biostatistics, Erasmus University Medical Center, the Netherlands\\
	$^2$Department of Epidemiology, Erasmus University Medical Center, the Netherlands\\
	$^3$Department of Child Psychiatry, Erasmus University Medical Center, the Netherlands\\
	$^4$Department of Psychology, Education and Child Studies, Erasmus University Rotterdam, the Netherlands\\}
	\vspace{10pt}
	\today \\
}
\begin{document}
\maketitle
\begin{abstract}
The experience sampling method (ESM) is a longitudinal research design where participants
report their thoughts, emotional states and behaviours multiple times a day. Our work is
motivated by such data collected by the \text{GrowIt!} app, which was released to investigate
daily emotions among adolescents during the COVID-19 pandemic. Current procedures to
analyse ESM data face various challenges. While standard statistical techniques may not scale
well to a high-dimensional setting, machine learning procedures can give biased results due to
selection bias introduced by missingness. In our motivating dataset, adolescents dropped out
due to previous strong feelings of negative emotions. Hence, the implied missing data are of
the missing-at-random type that standard machine learning procedures cannot accommodate. We
develop a novel neural network architecture that generalises mixed effects models to deep
learning to overcome these challenges. It allows semi-parametric and flexible modelling of
data's mean and correlation structure through fixed and random effects. For estimation, we use
an adaptation of variational auto-encoders and a Bayesian data augmentation algorithm. Through
this approach, the model can accommodate longitudinal outcomes following generic distributions,
scale well to high-dimensional settings and provide valid inference when data are
missing-at-random. We applied the Deep Generalised Mixed Model to the \text{GrowIt!} study and
various simulations. The results show potential for the Deep Generalised Mixed Model, yet suboptimal 
performance due to model instability.
\end{abstract}
\keywords{longitudinal data, missing data, mixed models, neural networks, variational auto-encoders}
\section{Introduction}
\label{s:intro}
Experience sampling method (ESM) data are becoming increasingly popular in predicting health
outcomes in participants and improving their daily life experiences. The ESM is an intensive
longitudinal research design that asks participants to report on their thoughts, emotional
states and behaviours at multiple time points a day over a sustained period. Through this
design, researchers capture important daily life health factors such as anxiety and stress.
Our work is motivated by ESM studies performed via the $\text{GrowIt!}$ app. During the
COVID-19 pandemic, the GrowIt! app was released to improve adolescents’ affective and cognitive well-being. 
The app consists of two components; firstly monitoring the emotions and behaviours of adolescents in daily life using the ESM. 
Secondly, offering daily challenges that are based on CBT and aimed at strengthening adaptive coping\supercite{growit, growit_1, growit_2, growit_3}. 
Since then, the $\text{GrowIt!}$ app has been further tested in a
randomised controlled trial as an intervention to study its beneficial effects on a population
of children and adolescents aged 10 to 18 and diagnosed with chronic somatic conditions\supercite{growit_ct}.\\
\indent A common statistical framework for analysing longitudinal data is mixed effects models
\supercite{MixedModel, MM}. The advantage of this method is that it can accommodate common
features of longitudinal data, such as unbalanced data, irregular time intervals and drop-out.
However, ESM data are also characterised by $\text{high-dimensional}$ complex correlation
structures. Although extensions have been made in previous research to make mixed models more
efficient for high-dimensional purposes\supercite{FAST_MM1, FAST_MM2, FAST_MM3}, standard mixed model
implementations may either be too rigid or too computationally demanding to capture such
high-dimensional structures properly. Mixed effect models also pose strict assumptions on
the correlation structure that are violated.\\
\indent An alternative approach to modeling intensive longitudinal data is to use machine learning
(ML) techniques designed to accommodate such high-dimensionality\supercite{ESM_ML}. In the ML literature, deep
learning (DL) is often used to analyse longitudinal data. For a general introduction to DL, we
refer readers to Prince (2023)\supercite{intro_DL}. A well-known DL architecture for longitudinal 
data is recurrent neural networks (RNNs), which were designed to process sequential data by
using recurrent connections\supercite{RNN},
though other architectures exist such as Temporal Convolutional Neural Networks\supercite{TCNN}
and more recently transformers\supercite{Transform}. However, a significant issue remains for the
analysis of longitudinal data using these DL techniques. Namely, the selection bias that is
introduced through missing data in longitudinal studies. \\
\indent Although various procedures to deal with missing data in DL have been studied previously
\supercite{RNN_MCAR1, RNN_MCAR2, RNN_MCAR3, RNN_MCAR4, RNN_MCAR5}, most of these methods such as
mask matrices and mean imputation provide valid inference only when the missing data mechanism
is missing-completely-at-random (MCAR), i.e. independent of any of the observed data.
Nonetheless, previous research has shown that ML techniques may be robust in the face of
missing data when the aim is prediction no matter the missing data mechanism\supercite{MARP1,
MARP3, MARP4}. In response, Catoire et al.\ (2026) %\supercite{MARPO} 
recently published a pre-print to introduce a new formal framework for prediction tasks that explains these
contradicting results between inference and prediction. However, the previous results and
framework all assume a standard regression scenario where there is either no missing data in
the outcome, or that the missing data in the outcome is non-informative. Yet, in longitudinal
studies, the missingness is more complex. There frequently is either intermittent missingness
in the outcome or drop-out over time. Furthermore, this drop-out usually depends on earlier
observed data, potentially introducing selection bias. For example, in our motivating dataset, youths may have missed measurements 
or completely stopped using the $\text{GrowIt!}$ app due to previous strong feelings of negative emotions\supercite{growit_miss}. In this scenario,
we have to appropriately account for the missing data if we want to accurately and unbiasedly
predict the future trajectories of individuals that dropped-out from our study. \\
\indent As far as we were able to find, only Che et al.\ (2018)\supercite{RNN_MAR} 
have researched a method
where a RNN leverages the information of patterns in the missing data to account for potential
selection bias and improve performance. However, their method makes two assumptions on the
missingness mechanisms. The method assumes that missing values decay back to some empirical
mean, and that the missingness mechanisms depends only on the time difference between
measurements. Both these assumptions may not hold in all scenarios. For example, a participant
who dropped out may have done so due to an abnormal state such as a major depressive disorder.
In this case, the patient may not return to their previous empirical mean. Furthermore,
missingness may also be related to other variables such as the previously measured outcomes. \\
\indent In this work, we overcome the issues described above for the analysis of intensive
longitudinal data by developing a novel neural network (NN) architecture that generalises
mixed effects models to DL, combining the advantages of both statistical techniques and
machine learning methods. We call this new framework the Deep Generalised Mixed Model (DGMM).
Previously, Mandel et al.\supercite{biometric_NN} generalised mixed models such that the fixed
effects were estimated using a NN and any remaining heterogeneity was modeled using random
intercepts. For the DGMM, we build upon their work by estimating both the fixed effects and
random effects using NNs, allowing for more flexible and semi-parametric modelling of both
within- and between-person differences. \\
\indent The remainder of the paper is organised as follows. In Section~\ref{s:model}, we formally
introduce the architecture, estimation procedure, missing data generalisation and forecasting
algorithm for the DGMM. In Section~\ref{s:application}, we apply the DGMM to the
$\text{GrowIt!}$ app studies. In Section~\ref{s:simulations}, we compare the performance of
the DGMM in various simulation studies to the performance of a RNN and a standard mixed
effects model. Finally, we conclude our findings and practical implications with a discussion
in Section~\ref{s:discussion}. \\
\section{Model Definition}
\label{s:model}
Let $\boldsymbol{Y_{i}} = (\boldsymbol{y}_{i,1}, \dots, \boldsymbol{y}_{i,K})$ denote
a matrix of $k = 1, \dots, K$ longitudinal outcomes for subjects $i = (1, \dots, n)$ of
subject- and outcome-specific sizes $n_{i,k}$ obtained at time points $t_{il,k} > 0$. We
also let $\boldsymbol{X}_{i}$ denote an $(n_{i,k} \times p)$ matrix of baseline covariates
excluding time. We model the correlations in the longitudinal data with the conditional
independence assumption given a vector of random effects $\boldsymbol{b}_i$ of length $u$,
i.e. the longitudinal measurements $\boldsymbol{Y_i}$ are assumed independent conditioning
on $\boldsymbol{b}_i$. We use a general model specification to accommodate longitudinal
outcomes following a generic distribution. Specifically,
\begin{equation} \label{eq:DGMM}
[\boldsymbol{y}_{i,k} \mid \boldsymbol{b}_i, \boldsymbol{X}_{i}, \boldsymbol{T}_{i,k};
\boldsymbol{\theta}] \sim \mathcal{F}_k \biggr( \mathcal{G}_k \Bigr\{
\boldsymbol{\mu}_1(\boldsymbol{X}_{i}, \boldsymbol{T}_{i,k}) +
\boldsymbol{\mu}_2(\boldsymbol{B}_{i}, \boldsymbol{T}_{i,k})\Bigr\}, \varsigma_k \biggr),
\end{equation}
in which
\begin{equation*}
\begin{aligned}
\relax[\boldsymbol{b}_i] &\sim \mathcal{N}(\boldsymbol{0}, \mathbf{I}), \\
\boldsymbol{\mu}_1(\boldsymbol{X}_{i}, \boldsymbol{T}_{i,k}) &= f(\boldsymbol{X}_{i},
\boldsymbol{T}_{i,k}; \boldsymbol{\beta}_1), \\
\boldsymbol{\mu}_2(\boldsymbol{B}_{i}, \boldsymbol{T}_{i,k}) &= g(\boldsymbol{B}_{i},
\boldsymbol{T}_{i,k}; \boldsymbol{\beta}_2).
\end{aligned}
\end{equation*}
The $k$-th longitudinal response $\boldsymbol{y}_{i,k}$ conditional on the random effects
$\boldsymbol{b}_i$ follows distribution $\mathcal{F}_k(\cdot)$ with mean defined by some
functions $g(\cdot)$ and $f(\cdot)$, parameterised by vectors $\boldsymbol{\beta_1}$ and
$\boldsymbol{\beta_2}$, and (potential) shape or dispersion parameter $\varsigma_k$. The
function $\mathcal{G}_{k}(\cdot)$ is a monotonic and differentiable link function and
$\boldsymbol{B}_{i}$ is the vector of $\boldsymbol{b}_i$, repeated to the length of $\boldsymbol{y}_{i,k}$. Lastly, $\boldsymbol{T}_{i,k}$ is a matrix of the vector of all timepoints $t_{il,k}$ and potential transformations of this vector.
The formulation in \eqref{eq:DGMM} presents a universal description of a generalised
multivariate linear mixed models, in which $\mathcal{F}_k(\cdot)$ can be a probability
distribution, such as a Gaussian, Student's \textit{t}, gamma, beta, binomial, Poisson or
negative binomial. Similar to a standard mixed model, the prediction equation is split up into
a fixed effect part, $\boldsymbol{\mu}_1(\cdot)$, to capture the population consistent trend,
and a random effect part, $\boldsymbol{\mu}_2(\cdot)$, to model remaining subject-specific
heterogeneity. However, compared to a mixed effects model, the DGMM assumes a (multivariate)
standard normal prior for the random effects $\boldsymbol{b}_i$ instead of a (multivariate)
normal distribution with mean(s) 0 and covariance matrix $\boldsymbol{D}$. \\
\indent To implement the model, we approximate the functions $g(\cdot)$ and $f(\cdot)$ through two
NNs. In particular,
\begin{equation*}
\begin{aligned}
f(\boldsymbol{X}_{i}, \boldsymbol{T}_{i,k}; \boldsymbol{\beta}_1) &= \text{NN}
\bigr\{ (\boldsymbol{X}_{i}, \boldsymbol{T}_{i,k}); \boldsymbol{\beta}_1,
\boldsymbol{\lambda}_{\beta_1} \bigr\}, \\
g(\boldsymbol{B}_{i}, \boldsymbol{T}_{i,k}; \boldsymbol{\beta}_2) &= \text{NN}
\bigr\{ (\boldsymbol{B}_{i}, \boldsymbol{T}_{i,k}); \boldsymbol{\beta}_2,
\boldsymbol{\lambda}_{\beta_2} \bigr\},
\end{aligned}
\end{equation*}
where we let $\text{NN}(\boldsymbol{X}; \boldsymbol{\beta}, \boldsymbol{\lambda})$ denote a
NN with input variables $\boldsymbol{X}$, parameters $\boldsymbol{\beta}$ and hyperparameters
$\boldsymbol{\lambda}$.
\subsection{Neural Network Architecture}
\label{s:arch}
\begin{figure}[!t]
    \centering
        \resizebox{1\textwidth}{!}{

            \begin{tikzpicture}[%transform canvas={scale=0.8},
                node distance=2.5cm and 2.5cm,
                every neuron/.style={
                    circle,
                    draw,
                    minimum size=1.2cm
                },
                latent neuron/.style={
                    circle,
                    draw,
                    fill = white,
                    minimum size=0.75cm
                },
                input neuron/.style={
                    draw=none,
                    minimum size=1.2cm
                },
                neuron missing/.style={
                    draw=none,
                    scale=1.2725,
                    text height=0.333cm,
                    execute at begin node=\color{black}$\vdots$
                }
            ]

            % ------------------- First Diagram -------------------

            \begin{scope}[xshift=-8cm]

            \begin{scope}[xshift=5cm]
            % Input Layer with t_{ij} at the top
            \node[input neuron] (I-1) at (0, 0) {$t_{ij}$};
            \node[input neuron] (I-2) at (0,-1) {$x_{i1}$};
            \node[neuron missing] (I-3) at (0,-2.5) {};
            \node[input neuron] (I-4) at (0,-4) {$x_{ip}$};

            % Hidden Layer L (First hidden layer)
            \node[every neuron] (H-1) at (3, 0.5) {$h^{(L)}_1$};
            \node[every neuron] (H-2) at (3,-1.5) {$h^{(L)}_2$};
            \node[neuron missing] (H-3) at (3,-3) {};
            \node[every neuron] (H-4) at (3,-4.5) {$h^{(L)}_{k_{h_L}}$};

            \foreach \i in {1,2,4}
                \foreach \j in {1,2,4}
                    \draw[->] (I-\i) -- (H-\j);

            \node at (5.5, -0.5) {\textbf{\dots}};
            \node at (5.5,-2) {\textbf{\dots}};
            \node at (5.5,-3.5) {\textbf{\dots}};

            \coordinate (B-1) at (6, 0.5);
            \coordinate (B-2) at (6, -1.5);
            \coordinate (B-3) at (6, -4.5);

            \foreach \i in {1,2,4}
                \foreach \j in {1,2,3}
                    \draw[-] (H-\i) -- ($(H-\i)!0.35!(B-\j)$);

            \coordinate (B-4) at (4, 0.5);
            \coordinate (B-5) at (4, -1.5);
            \coordinate (B-6) at (4, -4.5);

            % Second Hidden Layer
            \node[every neuron] (H1-1) at (8,0.5) {$h^{(1)}_1$};
            \node[every neuron] (H1-2) at (8,-1.5) {$h^{(1)}_2$};
            \node[neuron missing] (H1-3) at (8,-3) {};
            \node[every neuron] (H1-4) at (8,-4.5) {$h^{(1)}_{k_{h_1}}$};

            \foreach \i in {1,2,4}
                \foreach \j in {4,5,6}
                    \draw[->] ($(B-\j)!0.70!(H1-\i)$) -- (H1-\i);

            % Output Layer
            \node[every neuron] (O-0) at (10.5, -0.5) {$\mu_{1_{ij, 1}}$};
            \node[neuron missing] (O-1) at (10.5, -2) {};
            \node[every neuron] (O-2) at (10.5, -3.5) {$\mu_{1_{ij,K}}$};

            \foreach \i in {1,2,4}
            	\foreach \j in {0, 2}
                \draw[->] (H1-\i) -- (O-\j);

            \coordinate (Q1) at (11.5, -0.5);
            \draw[-] (O-0) -- (Q1);
            \coordinate (Q1-1) at (12, -3.5);
            \draw[-] (O-2) -- (Q1-1);

            \end{scope}

            % ------------------- Second Diagram -------------------

            \begin{scope}[yshift=-9cm]

            % Input Layer
            \node[input neuron] (I2-1) at (-3, 0) {$\boldsymbol{Y}_i$};
            \node[input neuron] (I2-2) at (-3,-2) {$\boldsymbol{t}_{i}$};
            \node[input neuron] (I2-4) at (-3,-4) {$\boldsymbol{X}_{i}$};

            % Hidden Layer L (First hidden layer)
            \node[every neuron] (H2-1) at (-0.5, 0.5) {$e^{(L)}_1$};
            \node[every neuron] (H2-2) at (-0.5,-1.5) {$e^{(L)}_2$};
            \node[neuron missing] (H2-3) at (-0.5,-3) {};
            \node[every neuron] (H2-4) at (-0.5,-4.5) {$e^{(L)}_{k_{e_L}}$};

            \foreach \i in {1,2,4}
                \foreach \j in {1,2,4}
                    \draw[->] (I2-\i) -- (H2-\j);

            \node at (1.25, -0.5) {\textbf{\dots}};
            \node at (1.25,-2) {\textbf{\dots}};
            \node at (1.25,-3.5) {\textbf{\dots}};

            \coordinate (B2-1) at (2, 0.5);
            \coordinate (B2-2) at (2, -1.5);
            \coordinate (B2-3) at (2, -4.5);

            \foreach \i in {1,2,4}
                \foreach \j in {1,2,3}
                    \draw[-] (H2-\i) -- ($(H2-\i)!0.35!(B2-\j)$);

            \coordinate (B2-4) at (0, 0.5);
            \coordinate (B2-5) at (0, -1.5);
            \coordinate (B2-6) at (0, -4.5);

            % Second Hidden Layer
            \node[every neuron] (H3-1) at (3,0.5) {$e^{(1)}_1$};
            \node[every neuron] (H3-2) at (3,-1.5) {$e^{(1)}_2$};
            \node[neuron missing] (H3-3) at (3,-3) {};
            \node[every neuron] (H3-4) at (3,-4.5) {$e^{(1)}_{k_{e_1}}$};

            \foreach \i in {1,2,4}
                \foreach \j in {4,5,6}
                    \draw[->] ($(B2-\j)!0.70!(H3-\i)$) -- (H3-\i);

            % Latent space: mu
            \node[latent neuron] (LM-1) at (5, 1.5) {$\mu_1$};
            \node[neuron missing] (LM-2) at (5, 0.5) {};
            \node[latent neuron] (LM-3) at (5, -0.5) {$\mu_u$};

            \foreach \i in {1,2,4}
              \foreach \j in {1, 3}
                   \draw[->] (H3-\i) -- (LM-\j);

            % Latent space: sigma
            \node[latent neuron] (LM-4) at (5, -3.5) {$\sigma_1$};
            \node[neuron missing] (LM-5) at (5, -4.5) {};
            \node[latent neuron] (LM-6) at (5, -5.5) {$\sigma_u$};

            \foreach \i in {1,2,4}
              \foreach \j in {4, 6}
                   \draw[->] (H3-\i) -- (LM-\j);

            % Samples
            \fill[gray!20] (7,-2) ellipse [x radius=0.75, y radius=1.85];
            \node[latent neuron] (Z-1) at (7, -1) {$z_1$};
            \node[neuron missing] (Z-2) at (7, -2) {};
            \node[latent neuron] (Z-3) at (7, -3) {$z_u$};
            \node[input neuron] (Z-4) at (7, -4.5) {$t_{ij}$};

            \draw[dashed] (LM-1) -- (Z-1);
            \draw[dashed] (LM-3) -- (Z-3);
            \draw[dashed] (LM-4) -- (Z-1);
            \draw[dashed] (LM-6) -- (Z-3);

            % Decoder
            \node[every neuron] (D1-1) at (9, 0.5) {$d^{(L)}_1$};
            \node[every neuron] (D1-2) at (9, -1.5) {$d^{(L)}_2$};
            \node[neuron missing] (D1-3) at (9,-3) {};
            \node[every neuron] (D1-4) at (9, -4.5) {$d^{(L)}_{k_{d_L}}$};

            \foreach \i in {1,3,4}
            	\foreach \j in {1,2,4}
            		\draw[->] (Z-\i) -- (D1-\j);

            \coordinate (B3-1) at (11.5, 0.5);
            \coordinate (B3-2) at (11.5, -1.5);
            \coordinate (B3-3) at (11.5, -4.5);

            \foreach \i in {1,2,4}
            	\foreach \j in {1,2,3}
            		\draw[-] (D1-\i) -- ($(D1-\i)!0.35!(B3-\j)$);

            \node at (11, -0.5) {\textbf{\dots}};
            \node at (11,-2) {\textbf{\dots}};
            \node at (11,-3.5) {\textbf{\dots}};

            \coordinate (B3-4) at (10, 0.5);
            \coordinate (B3-5) at (10, -1.5);
            \coordinate (B3-6) at (10, -4.5);

            \node[every neuron] (D2-1) at (13, 0.5) {$d^{(1)}_1$};
            \node[every neuron] (D2-2) at (13, -1.5) {$d^{(1)}_2$};
            \node[neuron missing] (D2-3) at (13,-3) {};
            \node[every neuron] (D2-4) at (13, -4.5) {$d^{(1)}_{k_{d_1}}$};

            \foreach \i in {1,2,4}
                \foreach \j in {4,5,6}
                    \draw[->] ($(B3-\j)!0.70!(D2-\i)$) -- (D2-\i);

            % Output Layer
            \node[every neuron] (O2-0) at (15.5, -0.5) {$\mu_{2_{ij, 1}}$};
            \node[neuron missing] (O2-1) at (15.5, -2) {};
            \node[every neuron] (O2-2) at (15.5, -3.5) {$\mu_{2_{ij,K}}$};

            \foreach \i in {1,2,4}
            	\foreach \j in {0, 2}
            	\draw[->] (D2-\i) -- (O2-\j);

            % conjoining of mu1 and mu2
            \coordinate (Q2) at (16.5, -0.5);
            \draw[-] (O2-0) -- (Q2);
            \coordinate (Q2-1) at (17, -3.5);
            \draw[-] (O2-2) -- (Q2-1);

            \coordinate (Q3) at (16.5, 4);
            \draw[-] (Q2) -- (Q3);
            \draw[-] (Q1) -- (Q3);

            \coordinate (Q3-1) at (17, 1);
            \draw[-] (Q1-1) -- (Q3-1);
            \draw[-] (Q2-1) -- (Q3-1);

            \node[every neuron] (Q4) at (19.5, 4) {$\hat{y}_{ij,1}$};
            \node[neuron missing] (Q4-1) at (19.5, 2.5) {};
            \node[every neuron] (Q4-2) at (19.5, 1) {$\hat{y}_{ij,K}$};

            \draw[->] (Q3) -- (Q4);
            \draw[->] (Q3-1) -- (Q4-2);

            \node (P1) at (16.25, 4) {$+$};
            \node (P2) at (16.75, 1) {$+$};
            \node (Link) at (17.75, 4.25) {$\mathcal{G}_{1}(\cdot)$};
            \node (Link-K) at (17.75, 1.25) {$\mathcal{G}_{K}(\cdot)$};
            \end{scope}

            \end{scope}
            \end{tikzpicture}
            }
    \caption{General Neural Network Architecture for the Deep Generalised Mixed Model for $K$
    longitudinal outcomes. The shaded part indicates that the units within are held constant
    within an individual.}
    \label{fig:arch}
\end{figure}
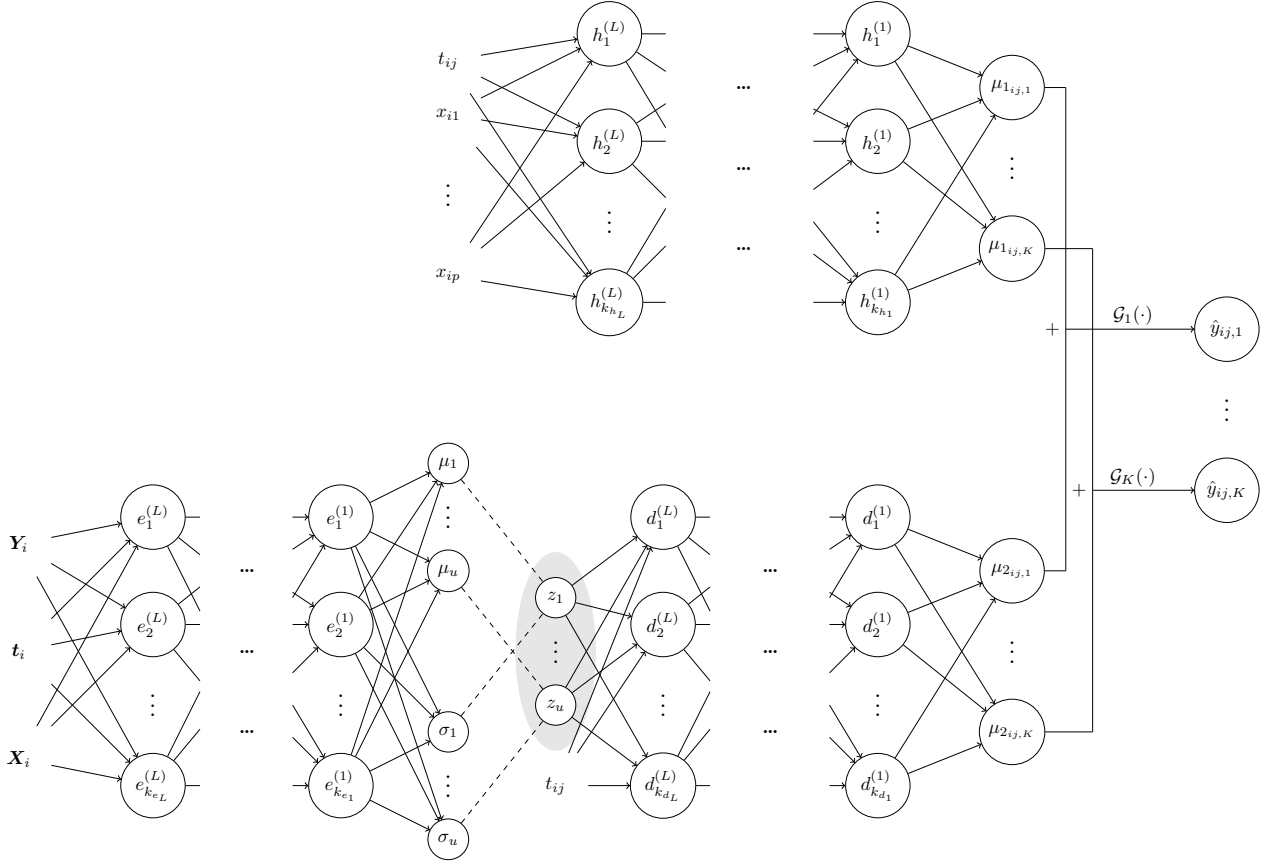

For notational simplicity, we assume $\boldsymbol{T}_{i,k}$ does not vary over the $K$
outcomes and denote it by $\boldsymbol{T}_i$. A general form for the architecture of the DGMM
for the $j$th measurement of the $i$th individual for the $K$ outcomes can be seen in
Figure~\ref{fig:arch}. The figure shows three NNs: $(1)$ $\boldsymbol{h} = \{h^{(L_h)},
\dots, h^{(1)}\}$, a standard feed-forward NN that models the fixed effects part
$\boldsymbol{\mu}_1(\cdot)$, $(2)$ $\boldsymbol{e} = \{e^{(L_e)}, \dots, e^{(1)}\}$, a NN
that encodes the data of individual $i$ into $u$ means $(\mu_1, \dots, \mu_u)$ and variances
$(\sigma_1, \dots, \sigma_u)$, and $(3)$ $\boldsymbol{d} = \{d^{(L_d)}, \dots, d^{(1)}\}$,
a NN that decodes samples $z_1, \dots, z_u$ from the encoded means and variances into a
subject-specific random effects part $\boldsymbol{\mu}_2(\cdot)$. These samples can be seen
as an approximation of the random effects $\boldsymbol{b}_i$ and are held constant within an
individual over all time points $\boldsymbol{T}_i$ and $K$ outcomes. Together, $\boldsymbol{e}$
and $\boldsymbol{d}$ create an architecture commonly known as a variational auto-encoder
(VAE), which is a generative DL method for simultaneously learning a deep latent variable
representation and probabilistic inference. For information regarding VAE, we refer readers to
Kingma and Welling (2019)\supercite{VAE}. \\
\indent The fixed effects network $\boldsymbol{h}$ has as input $t_{ij}$ and $\boldsymbol{X}_{i}=
(x_{i1}, \dots, x_{ip})$ with $L_h$ hidden layers of layer-specific lengths $k_{h_l} =
(k_{h_1}, \dots, k_{h_L})$ and multivariate outputs $\boldsymbol{\mu}_{1_{ij}} =
(\mu_{1_{ij,1}}, \dots, \mu_{1_{ij,K}})$. The outputs can be seen as a nonlinear function of
the inputs and structure of the neural network. Specifically, $\boldsymbol{X}_i$ and $t_{ij}$
enter the neural network through the first hidden layer $h^{(L)}$ with activation function
$g_{h_L}(\cdot)$ to produce the vector of length $k_{h_L}$:
\begin{equation*}
h^{(L)} = g_{h_L}( \omega^{h^{(L)}} \psi_i + \delta^{h^{(L)}} ),
\end{equation*}
where $\omega^{h^{(L)}}$ is a $k_{h_L} \times (p + 1)$ matrix of coefficients (often referred
to as weights in the DL literature), $\psi_i$ is shorthand notation for the combined vector
of $\boldsymbol{X}_{i}$ and $t_{ij}$ and $\delta^{h^{(L)}}$ is a vector of intercepts
(commonly known as biases in the DL literature) of length $k_{h_L}$. For the hidden layers in
$\boldsymbol{h}$ where $l = 2, 3, \dots L$, the output of the layer is defined by
\begin{equation*}
h^{(l - 1)} = g_{h_{l -1}}( \omega^{h^{(l - 1)}} h^{(l)} + \delta^{h^{(l - 1)}}),
\end{equation*}
where $\omega^{h^{(l-1)}}$ is a matrix of size $k_{h_{l - 1}} \times k_l$ and
$\delta^{h^{(l - 1)}}$ has length $k_{h_l}$ with layer-specific activation function
$g_{h_{l -1}}(\cdot)$. Lastly, the output layer of $\boldsymbol{h}$ is
\begin{equation*}
\boldsymbol{\mu}_{1_{ij}} = \omega^{h^{(1)}} h^{(1)} + \delta^{h^{(1)}},
\end{equation*}
of size $K$ and the identity activation function is used. \\
\indent As mentioned above, $\boldsymbol{e}$ and $\boldsymbol{d}$ take the form of a VAE. First, the
encoder $\boldsymbol{e}$ corresponds to the task of making a deep latent variable
representation of the data. For an individual $i$, all available data, i.e. $\boldsymbol{Y}_i$,
$\boldsymbol{T}_i$ and $\boldsymbol{X}_i$, enter the NN through the hidden layers
$\boldsymbol{e} = \{e^{(1)}, \dots, e^{(L)}\}$. Mathematically, the same happens within the
layers of $\boldsymbol{e}$ and $\boldsymbol{d}$ as it does in $\boldsymbol{h}$ with a
difference in hyperparameters, input and output. The encoder $\boldsymbol{e}$ outputs a layer
of $(2 \times u)$ units with the identity activation function that coincide with $u$ means
$\mu_1, \dots, \mu_u$ and $u$ log variances $\sigma_1, \dots, \sigma_u$, where $u$ is a
hyperparameter that defines the number of latent dimensions to encode the data into.
Specifically, the encoder of a VAE maps the data to a probability distribution in latent
space, as compared to directly estimating the latent dimensions as is done in a standard
auto-encoder, making the architecture less prone to overfitting and able to generate new data
\supercite{AE_to_VAE}. \\
\indent The second NN in the VAE is the decoder $\boldsymbol{d}$, which corresponds to the inference
task. From the $u$ means and log variances, we sample $\boldsymbol{z} = (z_1, \dots, z_u)$,
which are held fixed within an individual $i$. Together with measurement time $t_{ij}$, the
samples go through the layers of the decoder. As a result, the sampled $\boldsymbol{z}$ and
time point $t_{ij}$ are decoded into $\boldsymbol{\mu}_{2_{ij}} = (\mu_{2_{ij,1}}, \dots,
\mu_{2_{ij,K}})$, where the final layer $d^{(1)}$ again has a linear activation function. In
the last step, $\boldsymbol{\mu}_{1_{ij}}$ and $\boldsymbol{\mu}_{2_{ij}}$ are added together
element-wise. Each $k$-th addition is then put through link function $\mathcal{G}_k(\cdot)$
to obtain a prediction for the $k$-th longitudinal outcome for individual $i$ at time point
$t_{ij}$, i.e. $\hat{y}_{ij,k}$. \\
\subsection{Estimation}
\label{s:estimation}
Standard estimation of a set of model parameters $\boldsymbol{\theta}$ in a mixed model with
maximum likelihood estimation would involve calculation of the marginal likelihood function
\begin{equation*}
L(\boldsymbol{\theta}) = \prod_{i=1}^n p(\boldsymbol{Y}_i; \boldsymbol{\theta}) =
\prod_{i=1}^n \int p(\boldsymbol{Y}_i \mid \boldsymbol{b}_i; \boldsymbol{\theta}) \, p
(\boldsymbol{b}_i) \, \text{d}\boldsymbol{b}_i,
\end{equation*}
where the first term is the probability density of the conditional model for the outcome given
the random effects $\boldsymbol{b}_i$ and the second term is the probability density function
of the random effects. However, because $\text{log}\{p(\boldsymbol{Y}_i \mid \boldsymbol{b}_i;
\boldsymbol{\theta})\}$ of the DGMM is nonlinear in the random effects $\boldsymbol{b}_i$,
classical estimation by integrating out $\boldsymbol{b}_i$ to obtain the marginal likelihood
becomes intractable. This makes the model unfeasible, especially as the dimensionality of
$\boldsymbol{b}_i$ grows. \\
\indent To overcome this challenge in estimating $\boldsymbol{\theta}$, we use the variational
inference approach within a VAE to make the deep latent variable representation tractable.
Specifically, instead of estimating $p(\boldsymbol{b}_i)$ directly, we specify a tractable
variational distribution for the random effects $q(\boldsymbol{b}_i)$ from which we can
sample. Following a series of algebraic steps shown in Web Appendix~A, we establish that
maximising the marginal log-likelihood can be seen as equivalent to maximising the Evidence
Lower Bound\supercite{ELBO}:
\begin{equation*}
\tilde{\ell}(\boldsymbol{\theta}) = \sum_{i=1}^n \int \text{log} \bigr\{
p(\boldsymbol{Y}_i \mid \boldsymbol{b}_i; \boldsymbol{\theta}) \bigr\} \, q(\boldsymbol{b}_i)
\, \text{d}\boldsymbol{b}_i
- \int \text{log} \Bigr\{ \frac{q(\boldsymbol{b}_i)}{p(\boldsymbol{b}_i)} \Bigr\} \,
q(\boldsymbol{b}_i) \, \text{d}\boldsymbol{b}_i.
\end{equation*}
The first term in $\tilde{\ell}(\boldsymbol{\theta})$ can be approximated via Monte-Carlo
with
\begin{equation*}
\int \log \bigr\{ p(\boldsymbol{Y}_i \mid \boldsymbol{b}_i; \boldsymbol{\theta}) \bigr\}
q(\boldsymbol{b}_i) \, \text{d}\boldsymbol{b}_i
\approx \frac{1}{M} \sum_{m=1}^M \log \bigr\{
p(\boldsymbol{Y}_i \mid \boldsymbol{b}_i^{(m)}; \boldsymbol{\theta}) \bigr\},
\end{equation*}
where $\boldsymbol{b}_i^{(1)}, \dots, \boldsymbol{b}_i^{(M)}$ denote samples from the
variational distribution $q(\boldsymbol{b}_i)$. The second term in
$\tilde{\ell}(\boldsymbol{\theta})$ is the Kullback--Leibler (KL) divergence between the distributions $q(\boldsymbol{b}_i)$ and $p(\boldsymbol{b}_i)$\supercite{KL_Div}. When we set the variational distribution as a function of the data, i.e. $q(\boldsymbol{b}_i)
= p(\boldsymbol{b}_i \mid \boldsymbol{Y}_i; \boldsymbol{\theta})$, we obtain the result that
$\tilde{\ell}(\boldsymbol{\theta}) = \ell(\boldsymbol{\theta})$. This suggests that we should
specify the variational density $q(\boldsymbol{b}_i)$ such that the KL divergence between
$p(\boldsymbol{b}_i \mid \boldsymbol{Y}_i; \boldsymbol{\theta})$ and $q(\boldsymbol{b}_i)$ is
minimised. Linking these equations to the DGMM architecture in Figure~\ref{fig:arch},
$p(\boldsymbol{Y}_i \mid \boldsymbol{b}_i; \boldsymbol{\theta})$ is the addition of the NN
$\boldsymbol{d}$ and $\boldsymbol{h}$, and $p(\boldsymbol{b}_i \mid \boldsymbol{Y}_i;
\boldsymbol{\theta})$ are the estimated latent dimensions in $\boldsymbol{e}$. \\
\indent Using a variant of Bayesian central limit theorem\supercite{B_CLT} and under general regularity
conditions, we obtain that as the number of measurements within a person $n_{i,k}$ approaches
infinity, $p(\boldsymbol{b}_i \mid \boldsymbol{Y}_i; \boldsymbol{\theta})$ converges to a
multivariate normal distribution, i.e.
\begin{equation*}
[ \boldsymbol{b}_i \mid \boldsymbol{Y}_i; \boldsymbol{\theta} ] \sim \mathcal{N} \bigr\{
\boldsymbol{\mu}_b(\boldsymbol{Y}_i; \boldsymbol{\theta}),
\boldsymbol{\Sigma}_b(\boldsymbol{Y}_i; \boldsymbol{\theta}) \bigr\},
\end{equation*}
where
\begin{equation*}
\boldsymbol{\mu}_b(\boldsymbol{Y}_i; \boldsymbol{\theta}) = \operatorname*{arg\,max}_{\boldsymbol{b}}
\Big[ \log \bigr\{ p(\boldsymbol{Y}_i \mid \boldsymbol{b}; \boldsymbol{\theta}) \bigr\} +
\log \bigr\{ p(\boldsymbol{b}) \bigr\} \Big],
\end{equation*}
and
\begin{equation*}
\boldsymbol{\Sigma}_b(\boldsymbol{Y}_i; \boldsymbol{\theta}) =
\Bigg\{
- \frac{\partial^2}{\partial \boldsymbol{b}_i \partial \boldsymbol{b}_i^\top}
\Big[ \log \{ p(\boldsymbol{Y}_i \mid \boldsymbol{b}_i; \boldsymbol{\theta}) \} +
\log \bigr\{ p(\boldsymbol{b}_i) \bigr\} \Big] \bigg|_{\boldsymbol{b}_i = \boldsymbol{\mu}_b}
\Bigg\}^{-1}.
\end{equation*}
The normal approximation to the posterior is exact when $[\boldsymbol{Y}_i \mid
\boldsymbol{b}_i; \boldsymbol{\theta}]$ is Gaussian with a mean that is linear in
$\boldsymbol{b}_i$. We can also see $\boldsymbol{\mu}_b(\boldsymbol{Y}_i; \boldsymbol{\theta})$
and $\boldsymbol{\Sigma}_b(\boldsymbol{Y}_i; \boldsymbol{\theta})$ as nonlinear functions of
$\boldsymbol{H}_i = \{ \boldsymbol{Y}_i, \boldsymbol{T}_i, \boldsymbol{X}_i \}$. Hence, we
can approximate both using a NN architecture. In particular, we specify the variational
conditional distribution as a multivariate normal distribution, i.e.
\begin{align*}
\begin{cases}
[ \boldsymbol{b}_i \mid \boldsymbol{H}_i; \boldsymbol{\phi} ] \sim \mathcal{N} \bigr\{
\boldsymbol{\mu}_b(\boldsymbol{H}_i; \boldsymbol{\phi}),
\boldsymbol{\Sigma}_b(\boldsymbol{H}_i; \boldsymbol{\phi}) \bigr\}, \\
\bigr\{ \boldsymbol{\mu}_b(\boldsymbol{H}_i; \boldsymbol{\phi}),
\boldsymbol{\Sigma}_b(\boldsymbol{H}_i; \boldsymbol{\phi}) \bigr\} =
\text{NN}\{ \boldsymbol{H}_i; \boldsymbol{\phi}, \boldsymbol{\lambda}_\phi \}.
\end{cases}
\end{align*}
In Figure~\ref{fig:arch}, this is visualised as the encoder
$\boldsymbol{e}$. This structure estimates a $(n \times u)$ matrix of means and variances, and
the set of parameters $\boldsymbol{\phi}$ is assumed to be independent from $\boldsymbol{\theta}$.
Using this specification, we maximise
\begin{align*}
\tilde{\ell}(\boldsymbol{\theta}, \boldsymbol{\phi}) = \sum_{i=1}^n \biggr\{
&\int q(\boldsymbol{b}_i \mid \boldsymbol{H}_i; \boldsymbol{\phi}) \,
\text{log}\bigr\{ p(\boldsymbol{Y}_i \mid \boldsymbol{b}_i; \boldsymbol{\theta}) \bigr\} \,
\text{d}\boldsymbol{b}_i \, - \\
&\int q(\boldsymbol{b}_i \mid \boldsymbol{H}_i; \boldsymbol{\phi}) \, \text{log} \Bigr\{
\frac{q(\boldsymbol{b}_i \mid \boldsymbol{H}_i; \boldsymbol{\phi})}{p(\boldsymbol{b}_i)}
\Bigr\} \, \text{d}\boldsymbol{b}_i \biggr\}
\end{align*}
with respect to $\boldsymbol{\theta}$ and $\boldsymbol{\phi}$, where the first term is
approximated by
\begin{align*}
\int q(\boldsymbol{b}_i \mid \boldsymbol{H}_i; \boldsymbol{\phi}) \,
\text{log}\bigr\{ p(\boldsymbol{Y}_i \mid \boldsymbol{b}_i; \boldsymbol{\theta}) \bigr\} \,
\text{d}\boldsymbol{b}_i &\approx \frac{1}{M} \sum_{m=1}^M \text{log} \bigr\{
p(\boldsymbol{Y}_i \mid \boldsymbol{b}_i^{(m)}; \boldsymbol{\theta} \bigr\} \\
&= \frac{1}{M} \sum_{m=1}^M \sum_{j=1}^{n_i} \text{log} \bigr\{
p(y_{ij} \mid \boldsymbol{b}_i^{(m)}; \boldsymbol{\theta}) \bigr\},
\end{align*}
where $\boldsymbol{b}_i^{(1)}, \dots, \boldsymbol{b}_i^{(M)}$ now denote samples from
$q(\boldsymbol{b}_i \mid \boldsymbol{H}_i; \boldsymbol{\phi})$. The second term is now the
KL divergence between $\mathcal{N}(\boldsymbol{0}, \mathbf{I})$ and $\mathcal{N} \bigr\{
\boldsymbol{\mu}_b(\boldsymbol{H}_i; \boldsymbol{\phi}),
\boldsymbol{\Sigma}_b(\boldsymbol{H}_i; \boldsymbol{\phi}) \bigr\}$, which we can calculate
in closed form:
\begin{equation*}
\int q(\boldsymbol{b}_i \mid \boldsymbol{H}_i; \boldsymbol{\phi}) \, \text{log} \Bigr\{
\frac{q(\boldsymbol{b}_i \mid \boldsymbol{H}_i; \boldsymbol{\phi})}{p(\boldsymbol{b}_i)}
\Bigr\} \, \text{d}\boldsymbol{b}_i  = \frac{1}{2} \sum_{l=1}^k \mu_{il}^2 + \sigma_{il}^2 -
1 - \text{log}(\sigma_{il}^2),
\end{equation*}
where $\mu_{il}$ and $\sigma_{il}^2$ are the $l$-th components of the parameter vectors
$\boldsymbol{\mu}_b(\boldsymbol{H}_i; \boldsymbol{\phi})$ and
$\boldsymbol{\Sigma}_b(\boldsymbol{H}_i;\boldsymbol{\phi})$ for subject $i$, respectively.
Note that the NN structure can also be generalised to estimate a full variance-covariance
matrix by estimating the log Cholesky factor of this covariance matrix and changing the
formula for the KL divergence accordingly. This generalisation is presented in Web Appendix~B. \\
\indent All together, these steps allow the DGMM to estimate both fixed and random effects with NNs,
and let the model scale efficiently to high-dimensional data. \\
\subsection{Missing Data Generalisation}
\label{s:missing}
An implicit requirement for NNs is that all subjects are required to have the same input size.
However, a frequent feature of longitudinal data is that participants differ in the length of
the vectors $\boldsymbol{Y}_i$ and $\boldsymbol{T}_i$. Although this does not pose an issue
for the estimation of the decoder $\boldsymbol{d}$, it does complicate the encoder
$\boldsymbol{e}$. This is because $\boldsymbol{Y}_i$ and $\boldsymbol{T}_i$ are used as input
for $\boldsymbol{e}$. In existing NN structures for longitudinal data, this issue is usually
resolved using techniques such as zero-padding and mask matrices, where missing values are
removed from the estimation procedure. However, this method will not work for the encoder, as
we cannot mask missing values from contributing to the latent dimensions in the estimation of
the KL divergence. Furthermore, such a method is also not guaranteed to provide valid
inference under MAR without extensions as it does not leverage the information in the patterns
of missingness. \\
\indent To proceed, we assume that all subjects were supposed to provide longitudinal measurements at
some pre-specified time points. We redefine $\boldsymbol{T}_i$ into $\boldsymbol{T}_i^o$ and
$\boldsymbol{T}_i^m$ to respectively denote the time points of the observed longitudinal
measurements and the missing longitudinal measurements. Similarly, we redefine
$\boldsymbol{Y}_i$ into $\boldsymbol{Y}_i^o$ and $\boldsymbol{Y}_i^m$ and define
$\boldsymbol{H}_i^o = \{ \boldsymbol{Y}_i^o, \boldsymbol{T}_i^o, \boldsymbol{X}_i\}$ and
$\boldsymbol{H}_i^m = \{ \boldsymbol{Y}_i^m, \boldsymbol{T}_i^m, \boldsymbol{X}_i \}$. \\
\indent To properly train the encoder network $\boldsymbol{e}$ for $\boldsymbol{\mu}_b\bigr\{
(\boldsymbol{H}_i^o, \boldsymbol{H}_i^m); \boldsymbol{\phi}\bigr\}$ and
$\boldsymbol{\Sigma}_b \bigr\{ (\boldsymbol{H}_i^o, \boldsymbol{H}_i^m); \boldsymbol{\phi}
\bigr\}$ under missing data, we will impute the longitudinal measurements $\boldsymbol{Y}_i^m$
we were supposed to collect at $\boldsymbol{T}_i^m$. For the imputation process of
$\boldsymbol{Y}_i^m$, we borrow ideas from a stochastic approximation expectation maximisation
algorithm\supercite{StochEM} and combine this with a Robbins--Monro process\supercite{RM}. The
heuristic steps of the fitting algorithm for a total of $n_{\text{e}}$ number of epochs are
outlined in Algorithm~\ref{alg:fitting}.
\begin{algorithm}[t!]
\caption{DGMM Fitting Algorithm}
\label{alg:fitting}
\begin{algorithmic}[1]
	\State Define NN structures $\boldsymbol{h}$, $\boldsymbol{e}$ and $\boldsymbol{d}$.
	\State Obtain initial values for $\boldsymbol{Y}_i^{m,(0)}$ as part of
	$\boldsymbol{H}_i^{(0)} = \{ \boldsymbol{H}_i^o,\boldsymbol{Y}_i^{m,(0)},
	\boldsymbol{T}_i^{m} \}$.
\For{$v = 1$ \textbf{to} $n_{\text{e}}$}
		\State \fbox{Step 1:} Calculate $(\boldsymbol{\theta}^{(v)}, \boldsymbol{\phi}^{(v)})
		= \underset{\boldsymbol{\theta}, \boldsymbol{\phi}}{\arg\max} \, \bigr\{
		\tilde{\ell}_v(\boldsymbol{\theta}^{}, \boldsymbol{\phi}^{}; \boldsymbol{H}_i^{(v-1)}
		) \bigr\}$.
		\State \fbox{Step 2:} Simulate $\boldsymbol{Y}_i^{m,(v)}$ from
		$[\boldsymbol{Y}_i^m \mid \boldsymbol{Y}_i^o; \boldsymbol{\theta}^{(v)}]$.
		\State \fbox{Step 3:} Update $\boldsymbol{H}_i^{(v)} = \{
		\boldsymbol{H}_i^o,\boldsymbol{Y}_i^{m,(v)}, \boldsymbol{T}_i^{m} \}$.
	\EndFor
\end{algorithmic}
\end{algorithm}
First, we initialise the model with parameters $\boldsymbol{\theta}^{(0)}, \boldsymbol{\phi}^{(0)}$
and input values $\boldsymbol{H}_i^{(0)}$. Then, at the $v$-th iteration $(v = 1, \dots,
n_{\text{e}})$ in Step~1, we use a stochastic gradient descent algorithm (e.g.,\supercite{Adam}) to optimise $\tilde{\ell}_v(\boldsymbol{\theta}^{},
\boldsymbol{\phi}^{}; \boldsymbol{H}_i^{(v-1)})$ and obtain updated estimates of the
parameters $\boldsymbol{\theta}^{(v)}$ and $\boldsymbol{\phi}^{(v)}$. In Step~2, we use these
updated parameters to simulate a realisation of $\boldsymbol{Y}_i^{m, (v)}$ from the
corresponding posterior predictive distribution, i.e.
\begin{equation}
\label{eq:PPD}
p(\boldsymbol{Y}_i^{m,(v)} \mid \boldsymbol{Y}_i^o; \boldsymbol{\theta}^{(v)}) = \int
p(\boldsymbol{Y}_i^{m,(v)} \mid \boldsymbol{Y}_i^o; \boldsymbol{\theta}^{(v)})
p(\boldsymbol{b}_i \mid \boldsymbol{Y}_i^o ; \boldsymbol{\theta}^{(v)}) \,
\text{d}\boldsymbol{b}_i.
\end{equation}
We use a Random Walk Metropolis--Hastings (RWMH) algorithm to sample \eqref{eq:PPD}. The
exact inner steps performed in Step~2 and the
Robbins--Monro process are detailed in Web Appendix~C. Finally, in Step~3 we create $\boldsymbol{H}_i^{(v)}$ in which the missing values
$\boldsymbol{Y}_i^m$ are updated using the simulated $\boldsymbol{Y}_i^{m,(v)}$ from Step~2
for the following epoch.
\subsection{Dynamic Predictions and Prediction Error}
\label{s:dynpred}
Once we have estimated the DGMM in a dataset $\mathcal{D}_n = \{ \boldsymbol{y}_i,
\boldsymbol{X}_i, \boldsymbol{T}_i; i = 1, \dots, n\}$, we are interested in estimating
predictions for the longitudinal outcomes of a new subject $j$ from the same population.
Formally, let $\mathcal{Y}_j(t)$ denote the set of recorded measurements for the $K$ outcomes
of subject $j$ up to time $t$. We are interested in the predictions
\begin{equation*}
\hat{y}_{j,k}(u, t) = E \bigr\{ y_{j,k}(u) \mid \mathcal{Y}_j(t), \mathcal{D}_n \bigr\},
\quad u > t, \,\,\, t \geq 0.
\end{equation*}
Under an asymptotic Bayesian formulation of the posited model, we obtain
\begin{align*}
E &\bigr\{ y_{j,k}(u) \mid \mathcal{Y}_j(t), \mathcal{D}_n \bigr\} = \int E\bigr\{
y_{j,k}(u) \mid \mathcal{Y}_j(t); \boldsymbol{\theta} \bigr\} \, p(\boldsymbol{\theta} \mid
\mathcal{D}_n) \, \text{d}\boldsymbol{\theta} \\
&= \int \Biggr\{ \int E \bigr\{ y_{j,k}(u) \mid \boldsymbol{b}_j; \boldsymbol{\theta} \bigr\}
\, p(\boldsymbol{b}_j \mid \mathcal{Y}_j(t); \boldsymbol{\theta}) \, \text{d}\boldsymbol{b}_i
\Biggr\} \, p(\boldsymbol{\theta} \mid \mathcal{D}_n) \, \text{d}\boldsymbol{\theta} \\
&= \int \Biggl[ \int \mathcal{G}_k \Bigr\{ \boldsymbol{\mu}_1(\boldsymbol{X}_{j}, u;
\boldsymbol{\beta_1}) + \boldsymbol{\mu}_2(\boldsymbol{b}_{j}, u; \boldsymbol{\beta_2})\Bigr\}
\, p(\boldsymbol{b}_j \mid \mathcal{Y}_j(t); \boldsymbol{\theta}) \, \text{d}\boldsymbol{b}_i
\Biggl] \, p(\boldsymbol{\theta} \mid \mathcal{D}_n) \, \text{d}\boldsymbol{\theta}.
\end{align*}
We focus on point predictions and calculate $\hat{y}_{j,k}(u, t)$ using the estimated
parameters $\hat{\boldsymbol{\theta}} = \{ \hat{\boldsymbol{\beta_1}}, \hat{\boldsymbol{\beta_2}},
\hat{\boldsymbol{\varsigma}}\}$. The inner integral is approximated using Monte Carlo
sampling, i.e.
\begin{align*}
\int &\mathcal{G}_k \Bigr\{ \boldsymbol{\mu}_1(\boldsymbol{X}_{j}, u;
\hat{\boldsymbol{\beta_1}}) + \boldsymbol{\mu}_2(\boldsymbol{b}_{j}, u;
\hat{\boldsymbol{\beta_2}})\Bigr\} \, p(\boldsymbol{b}_j \mid \mathcal{Y}_j(t);
\hat{\boldsymbol{\theta}}) \, \text{d}\boldsymbol{b}_i \\ &\approx
\frac{1}{M} \sum_{m = 1}^M \mathcal{G}_k \Bigr\{ \boldsymbol{\mu}_1(\boldsymbol{X}_{j}, u;
\hat{\boldsymbol{\beta_1}}) + \boldsymbol{\mu}_2(\boldsymbol{b}_{j}^{(m)}, u;
\hat{\boldsymbol{\beta_2}})\Bigr\},
\end{align*}
where $(\boldsymbol{b}_j^{(1)}, \dots, \boldsymbol{b}_j^{(M)})$ denotes a sample from
$p(\boldsymbol{b}_i \mid \mathcal{Y}_j(t); \hat{\boldsymbol{\theta}})$ obtained using a RWMH
algorithm analogous to the one described in Web Appendix~C. \\
\indent When investigating the performance of the DGMM in the simulation studies, we use the dynamic
root mean square prediction error and dynamic bias, calculated by the formulas respectively
\begin{align*}
\text{RMSPE}(u, t) &= \sqrt{E \bigl\{ \bigl( y(u) - \hat{y}(u, t) \bigr)^2 \bigr\}}, \\
\text{bias}(u, t)  &= E \bigl\{ y^{*}(u) - \hat{y}(u, t) \bigr\}.
\end{align*}
where the expectation is taken over individuals with respect to the true data-generation
distribution with measurement error ($y(u)$) or without measurement error ($y^*(u)$). We use
an independent test set to obtain an objective estimate of the dynamic RMSE and bias.
Intuitively, we expect the $\text{RMSPE}(u, t)$ to increase as the time lag $| u - t |$
increases. \\
\section{The \text{GrowIt!} Study}
\label{s:application}
\begin{figure*}[t!]
\centering{
\includegraphics[scale=0.275]{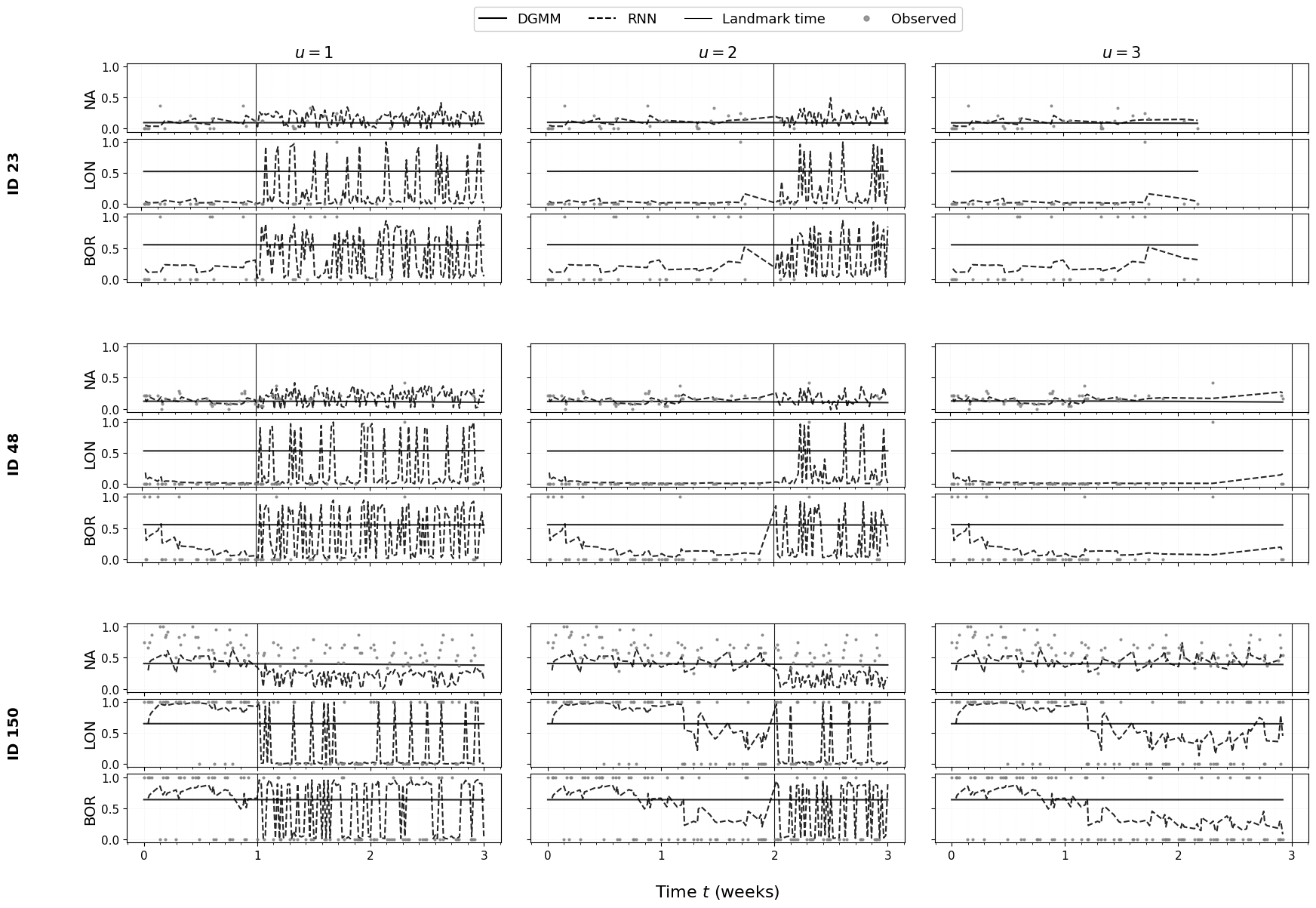}
}
\caption{Individualised dynamic predictions for three outcomes for three individuals in the
\text{GrowIt!} test set at three landmark times, denoted by $u$. NA = negative affect;
LON = loneliness; BOR = boredom; DGMM = Deep Generalised Mixed Model; RNN = Recurrent Neural
Network.}
\label{fig:Indv_Pred}
\end{figure*}

We return to our motivating example of the $\text{GrowIt!}$ studies during the COVID-19
pandemic to showcase the DGMM. The $\text{GrowIt!}$ app released three waves of studies
during the pandemic to research daily mood changes amongst young adults. We work with a subset
of the complete dataset that contains follow-up data of $1174$ participants and a total of
$123270$ measurement times, i.e. 105 measurements per participant. Among the participants,
only one provided data at all measurement occasions. Participants provided a median of 33
measurements (IQR = 40). In total $74831$ measurement times were missing, leaving us with
$48439$ observed measurements and a missingness proportion of approximately 0.61.
For the analysis and evaluation, we randomly split participants into a train, validation and
test dataset on a $7 : 1 : 2$ ratio. The train dataset was used to train the model, the
validation dataset was used for an early stopping criteria and the test set was used to
evaluate the performance of dynamic predictions. \\
\indent We modeled three ESM outcomes following various distributions: negative affect, a bounded
continuous outcome obtained by averaging four measurements on a $1$ to $7$ scale, and
loneliness and boredom, which were both measured on a scale from $1$ to $7$. We rescaled
negative affect to a range of 0 and 1 and dichotomised loneliness and boredom as lower than
$4$ or equal to/higher than $4$. Together, these outcomes describe an individual's negative
state in real-time. For the analysis, we included the baseline variable standardised age at
baseline. We specified time as first time since the participant started the study in weeks
with a range of 0 to 3. The specification of the DGMM was slightly different compared to how
we defined it in Section~\ref{s:model}. In particular, we made a more parsimonious network by
excluding the separate NN $\boldsymbol{h}$ (see Figure~\ref{fig:arch}) for the fixed effects.
Instead, we added the baseline information together with the time specifications to the decoder
$\boldsymbol{d}$. For the encoder $\boldsymbol{e}$, we used only $\boldsymbol{Y}_i$ as input
and $4$ hidden layers with $64$ units. We estimated the KL-divergence for 5 correlated latent
dimensions. For the decoder $\boldsymbol{d}$, we used $2$ hidden layers with $48$ units. We
chose a more shallow neural network for the decoder compared to the encoder to combat
posterior collapse, which is a common problem in VAE where the estimated latent dimensions
become equal to the prior distribution. A shallow decoder makes the model more inclined to
learn meaningful latent dimensions and not rely solely on baseline information and time to
reconstruct the data. For the same purpose, we also used a mix of cyclic
KL-annealing\supercite{VAE_PC_1} and $\beta$-VAE\supercite{VAE_PC_2}. We also added a warm-up
hyperparameter to aid convergence. Specifically, the warm-up denotes for how many epochs the
model should be trained before starting the imputation algorithm. This is because we found
that after only a few epochs of training, the error estimates were too large, leading to
inaccurate imputations and worse optimalisation. We employed an early stopping criteria to
improve generalisability of the model. In particular, the early stopping criteria minimised
the sum of the mean squared error for negative affect, and the log binomial loss for loneliness
and boredom in the validation data. Finally, we fitted a sequence-to-sequence multivariate RNN
with $3$ Long-Short-Term-Memory\supercite{LSTM} (LSTM) layers with $34$ units each for
comparison. \\
\indent We present the results of the multivariate analysis of the ESM outcomes in
Table~\ref{tab:results} and Figure~\ref{fig:Indv_Pred}. Table~\ref{tab:results} shows the
predictive accuracy of the DGMM and RNN on all observed data of the train, validation and
independent test dataset for the three outcomes. Interpreting these results, we see that the
DGMM showed poorer fit on the data compared to the RNN, especially on the binary outcomes
Loneliness and Boredom.
\begin{table*}[t!]
\centering
\begin{tabular}{ l | ccc | ccc | ccc }
\toprule
 & \multicolumn{3}{c}{Negative Affect} 
 & \multicolumn{3}{c}{Loneliness} 
 & \multicolumn{3}{c}{Boredom} \\
Model 
 & Train & Val & Test 
 & Train & Val & Test 
 & Train & Val & Test \\
\toprule

RNN 
 & 0.018 & 0.016 & 0.019 
 & 0.90 & 0.89 & 0.84 
 & 0.83 & 0.84 & 0.85 \\

DGMM 
 & 0.029 & 0.020 & 0.023
 & 0.76 & 0.81 & 0.86 
 & 0.70 & 0.79 & 0.80 \\

\midrule
\bottomrule
\end{tabular}
\caption{Accuracy metrics for the Deep Generative Mixed Effects Model and Recurrent Neural
Network across the datasets for the multivariate analysis of three outcomes. For Negative
Affect, the values denote the mean squared error. For Loneliness and Boredom, the values
denote the Area Under the Receiver Operating Characteristic Curve.}
\label{tab:results}
\bigskip
\end{table*}
Figure~\ref{fig:Indv_Pred} describes the individualised dynamic predicted trajectory for
three participants in the independent test set for landmark times $u = \{1, 2, 3\}$ in weeks.
From the observed data in the figure, we see that ID 150 experienced a prolonged period of
much higher negative affect, and more frequent loneliness and boredom than ID 23 and 48. Based
on this information that can be updated in real-time, a treating psychologist may look at the
trajectories and opt to intervene by sending a message to this individual and ask whether they
would like to have a consultation or provide a set of challenges on the app to aid their
affective state, similar to just-in-time adaptive interventions\supercite{jitai}. 
Preferably, such decisions would also be made on the predicted trajectory
past the observed data. However, we find that both the DGMM and RNN are unable to forecast an
individual's negative state well. While the DGMM collapsed back to a linear model that
provides a stable prediction past the landmark times, the RNN shows an overly flexible
trajectory which follows the observed data better, yet provides a very unstable forecast.
Furthermore, inspecting the forecasting of negative affect, we see that the DGMM keeps the
trend at the same level for ID 150 which follows the true data better than the RNN, which
predicts a large and inaccurate drop. This could be related to potential selection bias that
is introduced through the missing data in the outcome, which the DGMM takes into account. \\
\indent Summarising the results, we contend the unfavourable performance for the DGMM and RNN may be
caused by a shortcoming of information in the data. In particular, small sequential
correlations in the data, strong right skewed distributions for the three outcomes and the high
proportion of missing data make the data unsuitable for a prediction model. Instead clinical
decision making as described above should be based on observed data only. Experts could design
clinical cut-off values that would automate alerting a treating psychologist to save time and
resources. For example, a participant should be contacted if they have experienced bouts of
loneliness and boredom for more than three moments per day for two consecutive days combined
with a high negative affect score.
\section{Simulation Studies}
\label{s:simulations}
We conducted numerical studies to empirically compare the performance of the DGMM, a RNN
using LSTM cells and the true data generating model under various settings.
\subsection{Simulation Design}
\label{s:simdesign}
We designed our simulations to be a simplified version of longitudinal data. In particular,
we generated univariate continuous longitudinal data from a linear mixed model for $1200$
individuals with $20$ measurements each. The linear mixed model was specified to have
nonlinear trajectories over time using natural splines with three degrees of freedom. \\
\begin{figure*}[t!]
\centering{
\includegraphics[scale=0.55]{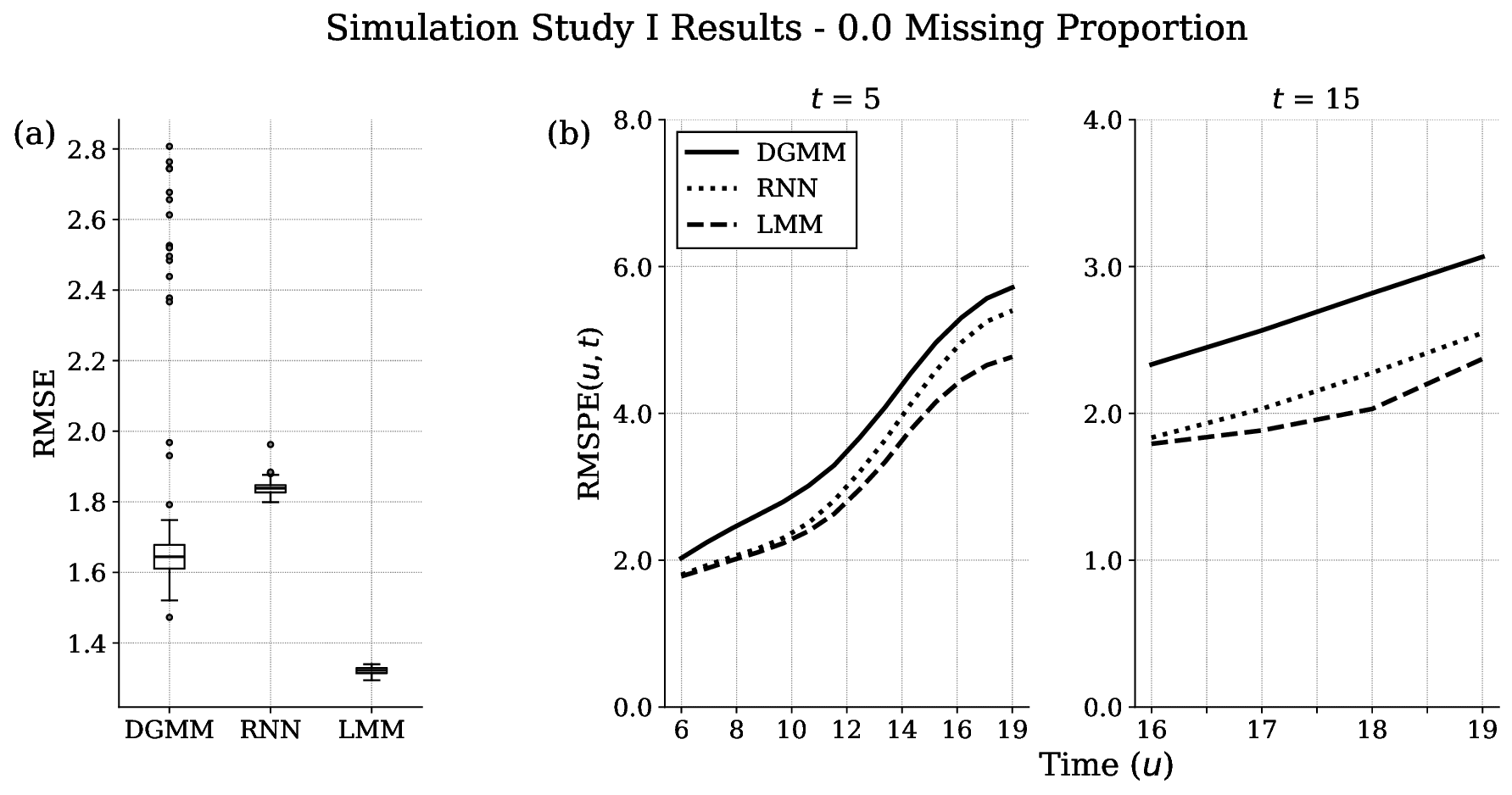}
}
\caption{Plot (a) on the left depicts the RMSE in the train data for the three tested methods.
Plots (b) on the right depict the average $\text{RMSPE}(u, t)$ for three methods at landmark
times $t = \{5, 15\}$ in the test data. DGMM = Deep Generalised Mixed Model; RNN = Recurrent
Neural Network; LMM = Linear Mixed Model.}
\label{fig:S1}
\end{figure*}
\indent After generating data, we apply a variety of missing data mechanisms for the different
simulation studies. Specifically, for Simulation Study~I, we assume no missing data and leave
the generated data as is. For Simulation Study~II, we assumed two types of MAR mechanisms. In
the first mechanism, drop-out was decided using a probabilistic scheme: at each measurement
time, the probability to have subsequent measurements missing from the study was estimated
using a logistic regression such that the higher someone's outcome value, the higher the
probability to drop-out. In the second mechanism, we specified a cut-off value such that if a
measurement was taken for an individual above this value, all subsequent measurements would be
missing. We varied the probabilities and cut-off value to have two different proportions of
missing data (approximately 0.10 and 0.30). In total, this led to a total of $5$ unique
simulation scenarios over the two simulation studies. We repeated each unique simulation
scenario 100 times. \\
\indent In each simulation scenario, we analysed the data in the following way. First, we split the
data into a training, validation and test dataset, dividing individuals on a $4 : 1 : 1$ ratio respectively.
Then, we train (a) the true model, i.e. a linear mixed model with natural splines, (b) the
DGMM and (c) a RNN with LSTM cells on the training dataset. For the LMM, the validation data
is included to train the LMM, whereas for the DGMM and RNN, we use the validation data for an 
early-stopping criteria. \\
\indent After model fitting, we estimated the $\text{RMSPE}(u, t)$ and $\text{bias}(u, t)$ for the
three methods for $t = \{5, 10, 15\}$ and $u = \{t + 1, \dots, 19\}$, which we visualised
in line-graphs. We also ascertained the MSE of the different methods in the training dataset,
which was visualised in box plots. Finally, we estimated the MSE of the imputed values that
were used as input in the encoder for the DGMM in the final fitting epoch and compared this
to the MSE had we imputed these values using the fitted true model.
\begin{figure*}[t!]
\centering{
\includegraphics[scale=0.60]{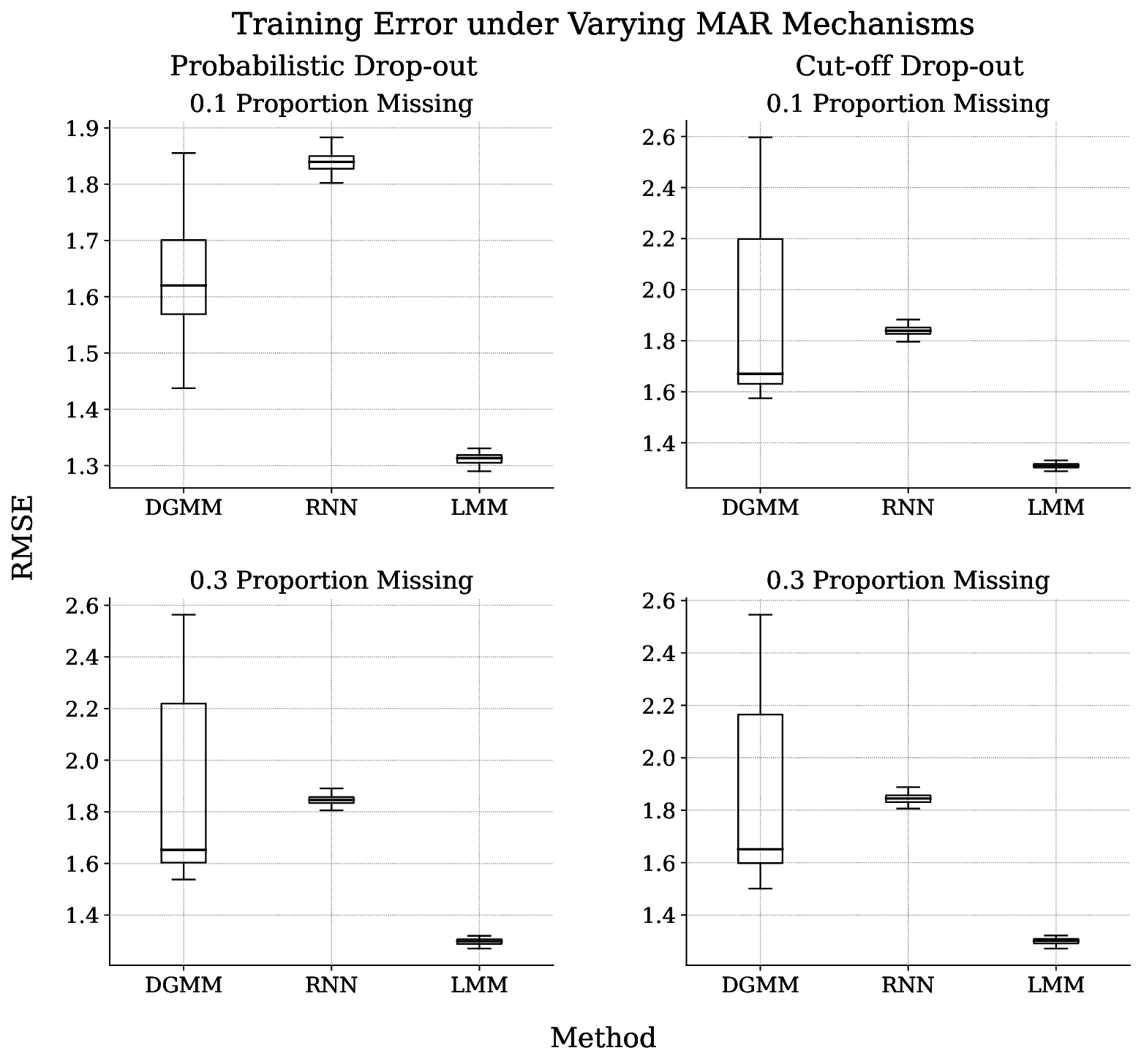}
}
\caption{Boxplots of the RMSE in the training data in Simulation Study~II for the three
methods under varying proportions missing data and MAR data generating mechanisms with outliers
removed. DGMM = Deep Generalised Mixed Model; RNN = Recurrent Neural Network; LMM = Linear
Mixed Model. For the full results with outliers, we refer readers to the Supplementary
Materials.}
\label{fig:S2_TE}
\end{figure*}
\subsection{Results}
\label{s:simresults}
\begin{figure*}[t!]
\centering{
\includegraphics[scale=0.60]{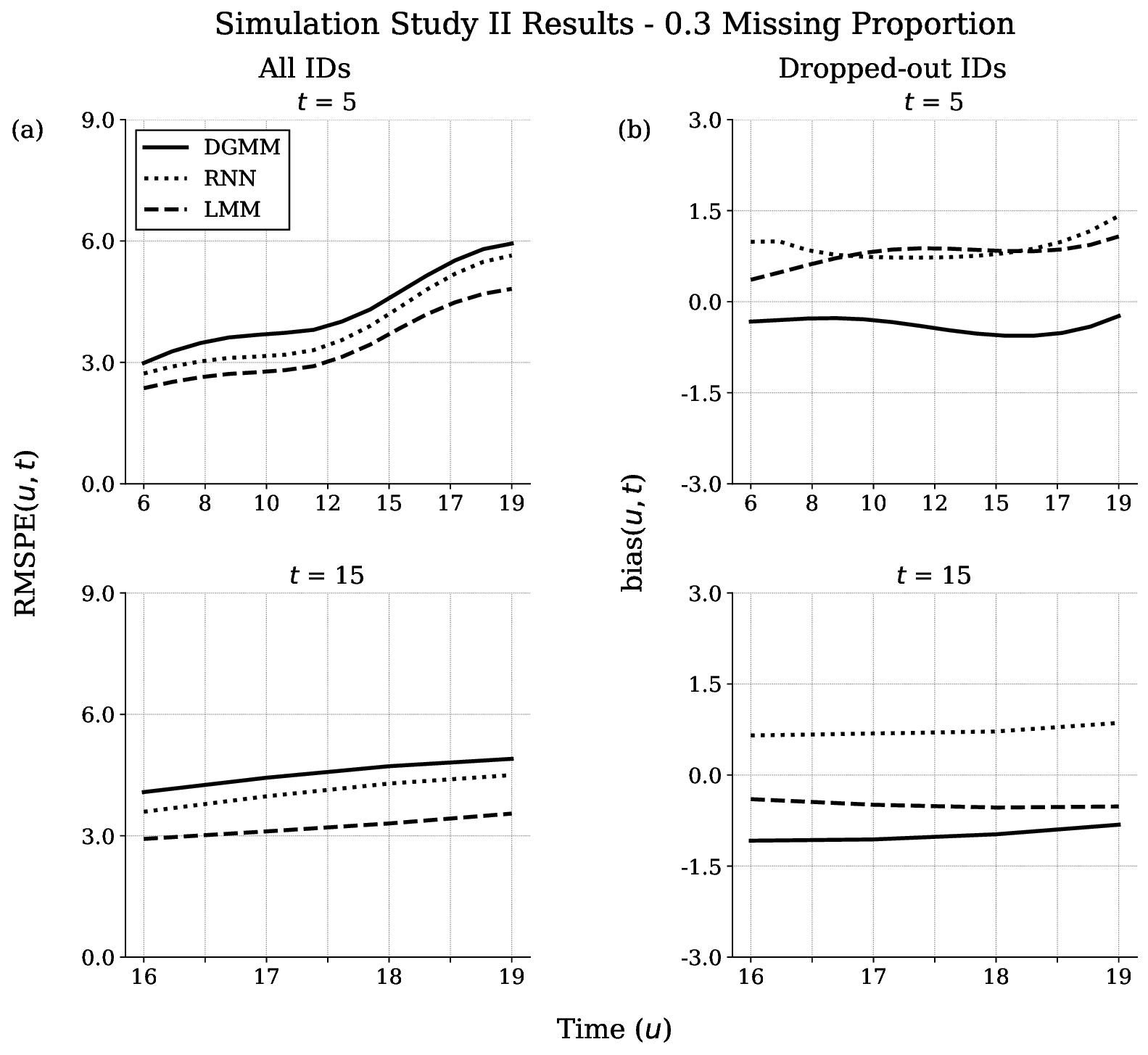}
}
\caption{Line graphs of the average $\text{RMSPE}(u, t)$ and $\text{bias}(u, t)$ for
$t = \{5, 15\}$ in an independent test set for the three methods under a proportion of 0.30
missing data and a probabilistic drop-out data generating mechanisms. The line graphs in the
left column are estimated over all simulated individuals, whereas the line graphs in the right
column are estimated only over the individuals that were simulated to drop-out according to the
missing data mechanism. DGMM = Deep Generalised Mixed Model; RNN = Recurrent Neural Network;
LMM = Linear Mixed Model.}
\label{fig:S2_FE}
\end{figure*}
Figure~\ref{fig:S1} presents the results for Simulation Study~I. Interpreting these results,
we find that under no missing data, the DGMM showcases good fit in the training data with slightly worse predictive performance compared to a RNN in an independent test set. \\
\indent In Simulation Study~II, where we introduce missing data into the design, we see that the
variation in the training process increased by a large margin (see Figure~\ref{fig:S2_TE}).
This is indicative that the DGMM suffers from model instability and may be susceptible to
small changes in hyperparameters and initialisation. We hypothesise this occurs due to the
variation introduced by varying number of observations per batch, the data imputation
algorithm, complemented by both the variation due to the variational inference in the DGMM
and inherent training variation in NNs introduced by batches, shuffling and initialisation.
This entails that there we see iterations where the DGMM is either underfitted or overfitted
in Simulation Study~II. \\
\begin{figure*}[t!]
\centering{
\includegraphics[scale=0.60]{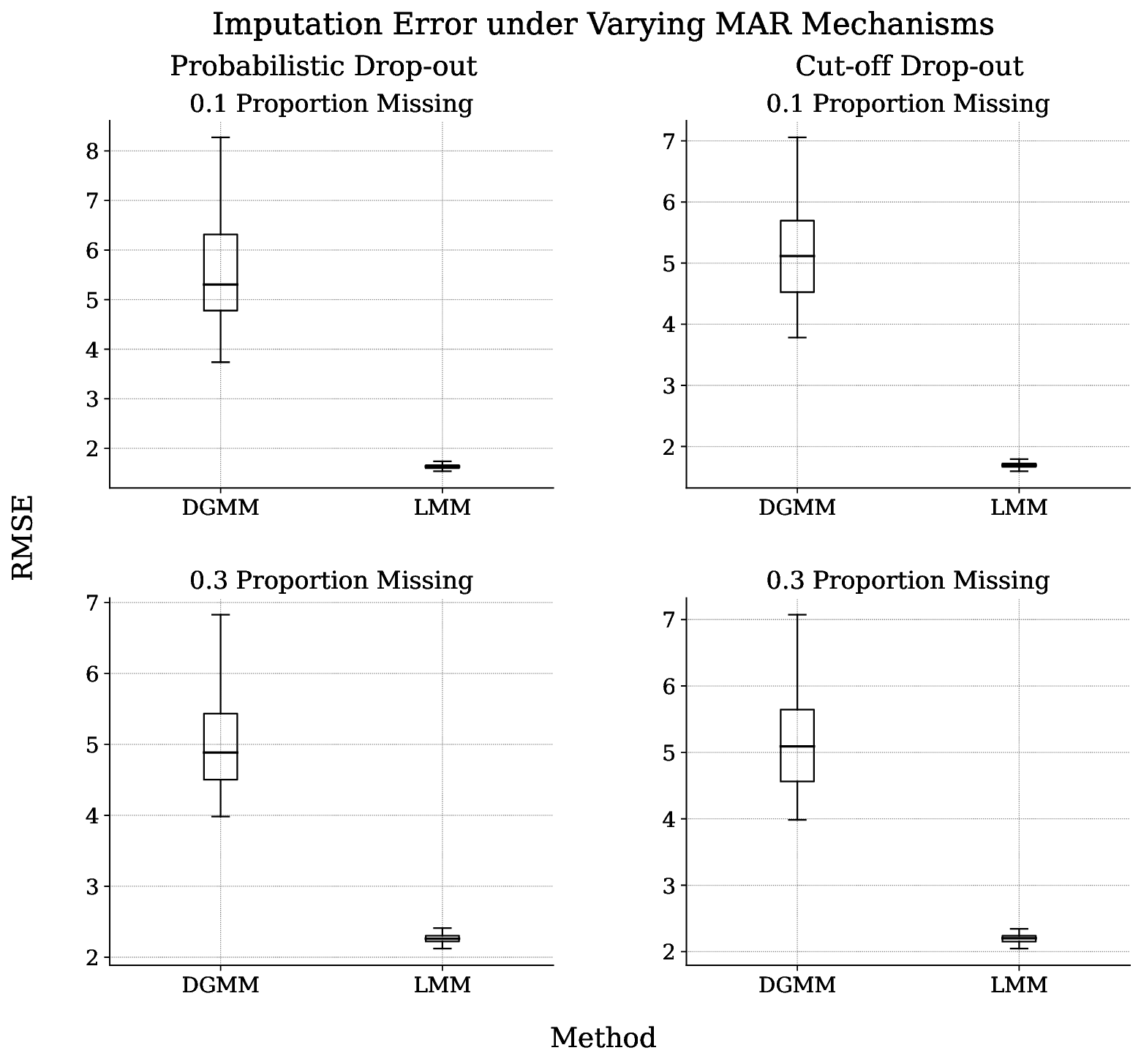}
}
\caption{RMSE of the data augmentation algorithm in Simulation Study~II for the DGMM under
varying proportions missing data and MAR data generating mechanisms with outliers removed.
DGMM = Deep Generalised Mixed Model; LMM = Linear Mixed Model. For the full results with
outliers, we refer readers to the Supplementary Materials.}
\label{fig:S2_IE}
\end{figure*}
\indent A consequence of the under- and overfitting of the DGMM is that there is large variation in
the predictive performance of the DGMM, which complicates interpretation of the results.
Nonetheless, we see that on average the performance of the DGMM is close to that of a RNN.
Across the simulation scenarios, we do find iterations where
the performance of the DGMM is higher than that of the RNN. In other words,
when the DGMM shows good convergence, the model may more properly leverage
information in the missing-at-random data mechanism and outperform a RNN, yet potentially fails to
do so on average due to model instability. 
Furthermore, it is worth noting that the differences in $\text{RMSPE}(u, t)$ are quite small when considering the scale of the outcome, which ranges from -20 to 20. In fact, from this perspective, all methods perform well. \\
\indent The benefit of the DGMM under MAR data is
further strengthened by the dynamic bias results in Simulation Study~II, which show that the
DGMM on average recovered the true trajectory more unbiasedly than a RNN. Due to the large
number of plots for Simulation Study~II, we depict only the results for the probabilistic
drop-out mechanism with a proportion of approximately 0.30 missing data in
Figure~\ref{fig:S2_FE}. For all plots of Simulation Study~II, we refer readers to
Web Appendix~D.  \\
\indent Lastly, we researched the accuracy of the data augmentation algorithm of the DGMM at the
final epoch and compare this to the accuracy had we imputed the values using the fitted data
generating model. Figure~\ref{fig:S2_IE} shows that the data augmentation algorithm can quite
accurately impute the missing values in the train data. However, we do again see a large
variance in this. \\
\section{Discussion}
\label{s:discussion}
In this work, we propose the DGMM, a DL generalisation of mixed effect models that
combines the benefits of statistical and machine learning techniques. The DGMM builds upon the
work of Mandel et al.\ (2021) by allowing both the fixed effects and random effects to be
estimated semi-parametrically using feed-forward NNs. As a result, the model becomes more
flexible and may more accurately model heterogeneity in hierarchical data. The model is
implemented using a VAE structure and through the use of a data augmentation algorithm should
provide valid inference and prediction under MAR data. \\
\indent We applied the DGMM for a multivariate analysis of the $\text{GrowIt!}$ ESM study to
showcase how the model may dynamically provide individualised predictions for outcomes
following varying distributions to aid decision making and compared this to a RNN, yet the
data did not lend itself to a prediction model. In our simulation studies, we found that
although the DGMM works properly and may outperform a RNN when data is MAR, especially as
more data is collected over time, the DGMM suffers from instability issues such as
hyperparameter sensitivity and posterior collapse, the latter of which is a known problem for
VAE. For future uses of the architecture, such instabilities should be taken into account
properly with techniques that are known to alleviate these (e.g.,
KL-annealing\supercite{VAE_PC_1}, $\beta$-VAE\supercite{VAE_PC_2} or guaranteeing identifiability
of the deep latent variable representation\supercite{VAE_PC_3}). \\
\indent Despite the subpar performance and issues we encountered in the application and simulations,
we believe that the DGMM can become a valuable architecture for the analysis of nonlinear
hierarchical data. For this article, we took a naive approach and used only standard layers in
feed-forward NN for the DGMM. However, future users could also employ other layers such as
LSTM layers or convolutional and pooling layers, commonly found in RNN and Temporal
Convolutional Neural Networks respectively, to make better use of the structure of the data.
Furthermore, although we focused solely on tabular intensive longitudinal data for the current
article, the DGMM could be extended to other data types of hierarchical data. For example,
medical imaging data where x-rays are taken of different body parts nested within individuals
with the exact structure of the NN adjusted to better model such data. \\
\indent A limitation to note for the DGMM is that the model trades interpretability for its
flexibility and semi-parametric estimation procedure. Future research could try to employ
explainable AI methods, such as SHAP values\supercite{shap}, to win back some of the lost
interpretability. Another part of the DGMM that we did not utilise is the deep latent variable
representation gained from the encoder. In our analysis, this was simply a byproduct. Future
studies may look into the explainability or potential of these latent dimensions in predicting
time-to-event outcomes by generalising the DGMM to a joint outcome setting as a shared
parameter model\supercite{jm} or in unsupervised machine learning tasks such as
clustering\supercite{VAE_clust_1, VAE_clust_2} (e.g., clustering the latent dimensions to find
different categories in the longitudinal trajectories).
\section*{Acknowledgements}
This research article was supported by Stress in Action. The research project `Stress in
Action' is financially supported by the Dutch Research Council and the Dutch Ministry of
Education, Culture and Science (NWO gravitation grant number 024.005.010).
\section*{Supporting Information}
Web Appendices referenced in Sections~\ref{s:estimation}, \ref{s:missing},
\ref{s:dynpred}, and~\ref{s:simresults} are available with this pre-print.
\section*{Data Availability Statement}
The data that support the findings of this study are available from the \text{GrowIt!} app COVID study.
Restrictions apply to the availability of these data, which were used under license for this
study. Data are available at \texttt{https://www.growitapp.nl} with the permission of the \text{GrowIt!} 
app research team.
\printbibliography
\end{document}